\documentclass[runningheads]{llncs}

\usepackage{eccv}

\usepackage{eccvabbrv}
\usepackage{graphicx}
\usepackage{tabularx}
\usepackage{booktabs}
\usepackage{multirow}
\usepackage{subcaption}
\usepackage{array}
\usepackage{makecell}
\usepackage{amsmath,amssymb}
\usepackage[accsupp]{axessibility}
\usepackage[hidelinks]{hyperref}
\usepackage[capitalize,nameinlink]{cleveref}

\newcommand{\ppc}{PPC}
\newcommand{\R}{\mathbb{R}}
\newcommand{\figplaceholder}[2]{%
  \fbox{\parbox[c][#2][c]{0.97\linewidth}{\centering #1}}%
}
\newcommand{\loadfig}[3]{%
  \IfFileExists{#1}{\includegraphics[#2]{#1}}{\figplaceholder{Missing figure: \texttt{\detokenize{#1}}}{#3}}%
}

\begin{document}

\title{Keep Your Friends Close, and the Right Neighbours Closer: Disaster-Conditioned Kernel-Regularized Graph Attention for Building Damage Classification}

\titlerunning{Kernel-Regularized Graph Attention for Damage Classification}

\author{Fuad Hasan\inst{1}\thanks{Corresponding author.} \and Chul Min Yeum\inst{1}}
\authorrunning{F. Hasan et al.}

\institute{
Computer Vision for Smart Structure (CViSS) Lab, University of Waterloo, Waterloo, Ontario, Canada\\
\email{fuad.hasan@uwaterloo.ca}
}

\maketitle

\begin{abstract}
Disaster damage is spatial: buildings rarely fail in isolation. Yet using spatial context for damage classification remains surprisingly underexplored, and many pipelines still rely primarily on per-building appearance cues even when the dominant uncertainty is spatially structured. Complicating matters, the \emph{right} neighbourhood is not the same across events. Floods, hurricanes, and wildfires can exhibit very different clustering behaviour, making spatial reasoning valuable but easy to misuse---naive context aggregation can improve visual coherence while oversmoothing boundaries or propagating structured errors. We study this tension on xBD (the dataset used in the xView2 challenge) in a controlled post-localization, classification-only setup: each building is represented by a \emph{pre/post combined} (\ppc) patch cropped from the provided polygons, and spatial context is modelled with GPS-derived building graphs. Our approach keeps local evidence ``close'' by preserving strong spatial relationships in disaster damage patterns, while bringing only the \emph{right} neighbours ``closer'' through a disaster-type-conditioned graph model that injects a learnable multi-scale spatial kernel prior into attention, allowing the effective neighbourhood scale to adapt across disaster types rather than being learned as a single global smoothing rule. To discourage coherence-by-smoothing, we add a residual de-correlation loss that penalizes positive Moran's~I in prediction residuals. We evaluate the method with (i) an xView2 holdout external-reference comparison under fixed building instances and (ii) zero-shot transfer evaluations that stress-test generalization under event and dataset shift: leave-one-event-out (LOEO) on xBD and cross-dataset transfer from xBD to Ida-BD. The model improves macro-F1 and substantially reduces residual spatial autocorrelation under zero-shot event shift, indicating better use of spatial context rather than naive smoothing and enabling more reliable transfer to unseen events within known disaster types.
\end{abstract}

\section{Introduction}
\label{sec:intro}
Rapid post-disaster building damage assessment from remote sensing imagery is a core operational task for emergency response: it supports triage, resource allocation, and recovery planning when time matters most. The xBD dataset, released for the xView2 challenge, made this problem tractable at scale by providing paired pre/post satellite imagery, building polygons, and per-building damage labels across multiple disaster types and regions \cite{gupta2019creatingxbd,gupta2019xbd,xview2dataset}. Yet two issues still limit real-world reliability: \textbf{cross-event generalization} and \textbf{spatially structured error}. Models that classify buildings independently from image patches often overfit event-specific appearance (roof materials, illumination, sensor differences), which gives rise to the errors or residuals sharing spatial clusters. Spatial context-aware models can improve local consistency but may blur boundaries and also propagate mistakes through neighbourhoods \cite{weber2020building,shen2022bdanet,deng2022improvedunet,velivckovic2018gat,brody2022gatv2}.

The key tension is that disaster damage is spatial---a behaviour famously coined as Tobler's First Law: "Everything is related to everything else, but near things are more related than distant things" \cite{tobler1970computer}; but, this spatial behaviour is not fixed \cite{cressie1993statistics}. In other words, \emph{nearby} is not a universal parameter: floods, hurricanes, wildfires, and earthquakes produce different neighbourhood-scale damage patterns, so a single notion of ``context'' is usually too rigid. In practice, this makes spatial reasoning both valuable and risky. If the model brings in the wrong neighbours, it can become visually smooth but systematically wrong---coherent maps that hide structured mistakes.

A key observation behind our method is that the strength and scale of spatial clustering vary substantially across events and disaster types. \Cref{tab:dataset_spatial_rollup} summarizes a compact event roll-up (our exact roll-up categorization is specified in Sec.~\ref{subsec:impl}) of ground-truth Moran's~I computed on xBD event tiles (using building severity labels mapped to \(\{0,1,2,3\}\)) for rollups with enough valid buildings. Hurricanes and wildfires show substantially stronger average spatial autocorrelation than floods, motivating type-adaptive context: when correlation extends further (as is often the case in some disaster types), more distant neighbours can be informative; when it does not, they should not be ``pulled closer'' by the model.

\begin{table}[t]
\centering
\small
\setlength{\tabcolsep}{4pt}
\caption{xBD spatial summary by disaster rollup. Moran's~I statistics are computed over valid train/test tiles.}
\label{tab:dataset_spatial_rollup}
\resizebox{\linewidth}{!}{%
\begin{tabular}{lccccc}
\toprule
\textbf{Rollup} & \textbf{Tiles (Train/Test)} & \textbf{Moran's~I mean$\pm$std (Train/Test)} & \textbf{Median Moran's~I (Train/Test)} & \textbf{Frac.\ $p<0.05$ (Train/Test)} & \textbf{Buildings/tile (Train/Test)} \\
\midrule
earthquake & 121 / 38  & 0.168 $\pm$ 0.222 / 0.356 $\pm$ 0.439 & 0.013 / 0.291 & 0.423 / 0.750 & 324.4 / 366.6 \\
flood      & 898 / 80  & 0.364 $\pm$ 0.482 / 0.195 $\pm$ 0.522 & 0.339 / 0.071 & 0.453 / 0.090 & 75.6 / 46.7 \\
hurricane  & 1219 / 387 & 0.306 $\pm$ 0.601 / 0.302 $\pm$ 0.589 & 0.226 / 0.203 & 0.233 / 0.260 & 72.5 / 75.0 \\
tornado    & 719 / 93   & 0.437 $\pm$ 0.317 / 0.320 $\pm$ 0.533 & 0.313 / 0.249 & 0.376 / 0.433 & 93.2 / 89.5 \\
tsunami    & 261 / 42  & 0.490 $\pm$ 0.407 / 0.484 $\pm$ 0.507 & 0.511 / 0.568 & 0.585 / 0.625 & 210.9 / 256.0 \\
volcano    & 309 / 75   & 0.593 $\pm$ 0.369 / 0.398 $\pm$ 0.481 & 0.339 / 0.340 & 0.273 / 0.129 & 23.2 / 15.0 \\
wildfire   & 5641 / 381 & 0.330 $\pm$ 0.625 / 0.568 $\pm$ 0.627 & 0.174 / 0.444 & 0.136 / 0.277 & 29.4 / 40.5 \\
\bottomrule
\end{tabular}%
}
\end{table}

This paper is a controlled post-localization study: given candidate building instances, we isolate whether relational spatial context improves building-level damage classification under event shift. We avoid detection and segmentation in the experimental loop and use the xBD polygons directly: each node is a \ppc\ building patch (pre/post crop), and graph edges are formed from GPS-derived building centroids from their polygons. This keeps the question focused: \emph{can we use neighbourhood context in a way that transfers across events---bringing in the "right" neighbours---without oversmoothing?}

Our answer is a \textbf{disaster-type-conditioned, multi-scale kernel prior} for graph attention, combined with a \textbf{residual de-correlation loss}. The kernel prior injects an interpretable, geostatistically motivated distance bias into attention and allows the effective neighbourhood scale to vary by disaster type, so the model can remain anchored to the per-building patch evidence while selectively amplifying informative neighbours. The residual term penalizes positive Moran's~I of prediction residuals \cite{moran1950notes,cressie1993statistics}, explicitly discouraging clustered errors. Together, these components push the model toward using spatial context as a predictive signal rather than as a smoothing heuristic.

More generally, the method can be viewed as a \emph{conditional context prior} for graph attention, consistent with relational-inductive-bias views of graph networks \cite{battaglia2018relational}: the effective neighbourhood scale is conditioned on a global/domain state rather than fixed globally. In this sense, the design is not an arbitrary distance embedding, but a monotone multi-scale log-prior attention mechanism paired with a residual spatial diagnostic; the same abstraction is relevant to structured vision problems where interaction strength varies with scene, geography, sensor, season, or event condition.

We evaluate the method in three complementary settings. First, the \textbf{xView2} holdout provides an external reference scale under the standard challenge setting \cite{xview2challenge}. Second, we use a harder \textbf{zero-shot leave-one-event-out (LOEO)} protocol on \textbf{xBD} to stress-test cross-event transfer and measure residual spatial structure, holding out one event at a time while providing its disaster type as metadata. Third, we test \textbf{zero-shot cross-dataset transfer} by training on xBD and evaluating directly on Ida-BD with a provided disaster-type token. Thus, xView2 contextualizes damage-classification scale, while xBD LOEO and xBD$\rightarrow$Ida-BD are the primary stress tests for context modelling under event and dataset shift.

The contributions are:
\begin{itemize}

    \item A \textbf{disaster-type-conditioned multi-scale (DCMS) kernel} parameterization that adapts correlation length scales across disaster types.
    \item A \textbf{kernel-regularized attention} layer for building damage classification that adds the disaster-type-conditioned spatial prior to GAT-style attention logits.
    \item A \textbf{residual spatial regularizer} based on Moran's~I, which explicitly discourages spatially clustered prediction errors.
    \item A \textbf{zero-shot transfer evaluation} that stress-tests context modelling under (i) xBD leave-one-event-out event shift and (ii) xBD$\rightarrow$Ida-BD cross-dataset shift.
\end{itemize}

\section{Related Work}
\label{sec:related}
\paragraph{Building damage assessment on xBD and xView2.}
The xBD dataset, introduced with the xView2 challenge, established large-scale, bitemporal satellite imagery for building damage assessment with polygon-level annotations and four damage categories \cite{gupta2019creatingxbd,gupta2019xbd}. Many strong methods use two-stage pipelines (localization + classification) or bitemporal fusion networks that explicitly compare pre/post imagery \cite{weber2020building,shen2022bdanet,deng2022improvedunet}. Recent dataset efforts such as Bright broaden building-damage assessment toward multimodal optical/SAR, all-weather EO benchmarks \cite{bright2025dataset}. xFBD further studies focused building damage assessment and object-level separability, including settings where surrounding context may bias classification \cite{melamed2023xfbd}. This is complementary to our motivation: spatial context should not be forced when labels are weakly clustered, and a conditional kernel prior can reduce neighbourhood influence in such regimes rather than smoothing indiscriminately. These methods are effective, but improvements can be entangled with detector quality, segmentation quality, and fusion design. Our setup removes those factors by using the provided polygons and focusing on building-level classification and contextual reasoning.

\paragraph{Graph-based context modelling.}
Graph neural networks offer a natural way to reason over nearby buildings. GraphSAGE \cite{hamilton2017inductive}, GCN \cite{kipf2017semi}, and GAT/GATv2 \cite{velivckovic2018gat,brody2022gatv2} provide increasingly expressive message passing, but standard attention is unconstrained and may learn unstable or over-smoothed spatial behaviour under event shifts. Prior work used vanilla graph attention (which we refer to as vanilla GAT) for spatially aware building damage assessment on UAV imagery and observed both benefits and oversmoothing-like failure modes in boundary-heavy scenes \cite{hasan2025uavgat}. This motivates adding an explicit spatial prior and a residual spatial diagnostic to the learning objective.

\paragraph{Geostatistics and spatial diagnostics.}
The idea that nearby observations tend to be more related (Tobler's first law) is foundational in spatial analysis \cite{tobler1970computer}. Moran's~I and Geary's~C remain standard diagnostics for spatial autocorrelation \cite{moran1950notes,geary1954contiguity,cressie1993statistics}. We borrow this lens for model design: instead of using spatial statistics only for post-hoc analysis, we incorporate a learnable spatial kernel into attention and explicitly penalize residual spatial autocorrelation during training.

\section{Method}
\label{sec:method}
\begin{figure}[t]
    \centering
    \loadfig{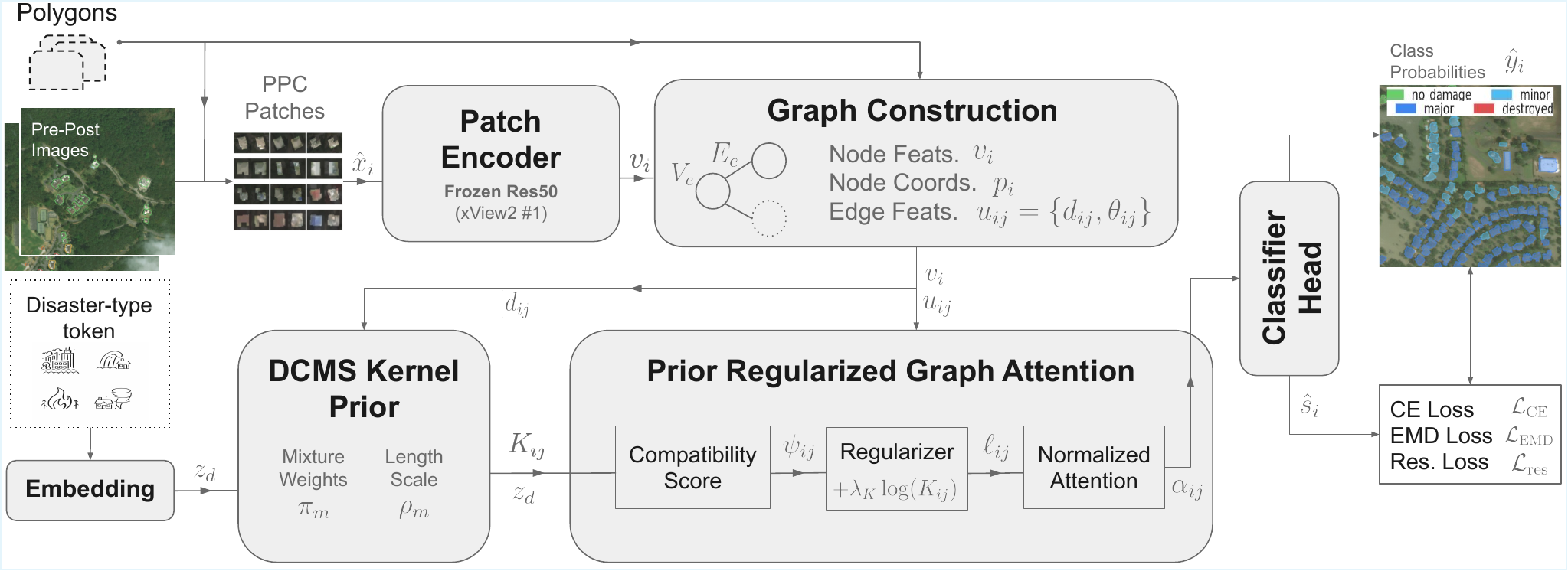}{width=0.98\linewidth}{2.1in}
  \caption{Method overview. \ppc\ patches form graphs over fixed building instances; a disaster-conditioned DCMS kernel regularizes graph-attention logits, and training combines classification/ordinal losses with residual Moran de-correlation.}
    \label{fig:method_overview}
\end{figure}

\Cref{fig:method_overview} gives the full pipeline. We first define a classification-only xBD task setup using \ppc\ building patches and GPS-derived graphs, then encode patches, run graph reasoning with disaster-type-conditioned kernel-regularized attention, and finally optimize both classification loss and a residual Moran objective.

\subsection{Task Setup and Graph Construction}
\label{subsec:task}
We use the xBD dataset \cite{gupta2019xbd,gupta2019creatingxbd,xview2dataset}, released for the xView2 challenge, and formulate a controlled post-localization, classification-only building damage task. Rather than treating xBD as a detection/segmentation task, we directly crop a \ppc\ building patch for each annotated polygon and classify damage at the building level. Although experiments use xBD polygons to isolate context modelling, the graph module itself only requires candidate building instances and centroids; in deployment these could come from detector/segmenter outputs or existing footprints (e.g., OSM/cadastre), with localization noise affecting crop quality, node dropout, and centroid jitter.

We process xBD as a collection of \emph{per-event} attributed graphs. Events are disjoint and we do not assume any persistent building identity across events; within each event \(e\), we simply index the \(N_e\) building polygons as \(i\in\{1,\dots,N_e\}\). For each building polygon, we compute a tight bounding box around the polygon vertices, add a small margin, crop the pre/post RGB regions, and resize to a fixed patch size. We refer to the resulting tensor as a \textbf{pre/post combined (\ppc) building patch}. In implementation this is a stacked bitemporal tensor, but we use the term \emph{patch} throughout for consistency.

The xBD annotations also include geographic coordinates. For each polygon \(i\) in event \(e\), we compute its centroid and project it to a local metric coordinate system (UTM per event/AOI), yielding a fixed position \(p_i\in\R^2\). We then build a \(k\)-nearest-neighbour (kNN) graph with node set \(V_e=\{1,\dots,N_e\}\) and directed edges \(E_e=\{(i,j): j\in\mathrm{kNN}(i)\}\). Each edge is derived from the fixed geometry and stores fixed edge features \(u_{ij}\) (concatenated distance \(d_{ij}=\|p_i-p_j\|_2\) and a relative direction encoding, \(\theta_{ij}\), which is a normalized displacement). We denote the resulting per-event attributed graph as
\begin{equation}
G_e=\Big(V_e,E_e,\{v_i\}_{i\in V_e},\{p_i\}_{i\in V_e},\{u_{ij}\}_{(i,j)\in E_e}\Big),
\end{equation}
where \(p_i\) and \(u_{ij}\) define the \emph{static} geometric scaffold, while \(\{v_i\}\) are learned node representations updated by the network. Importantly, the kNN \emph{topology} \((V_e,E_e)\) and geometric attributes \(\{p_i\}\), \(\{u_{ij}\}\) are fixed once constructed; graph attention does not add/remove edges, but learns \emph{edge weights} over this fixed neighbourhood during message passing.

\subsection{Patch Encoder}
\label{subsec:encoder}
For each building \(i\) in event \(e\), a \ppc\ patch \(x_i\) is fed to a pretrained ResNet-50 encoder \cite{he2016resnet} to produce an embedding \(v_i^{(0)}\in\R^d\). This encoder family follows the xView2 first-place-derived bitemporal fusion baseline \cite{weber2020building} and serves as the \emph{patch-only} baseline when used without graph context in later parts of the paper. We also concatenate simple polygon metadata (area and perimeter) to the embedding before graph reasoning. In contrast to the fixed coordinates \(p_i\), the node features are updated by message passing layers, yielding \(v_i^{(\ell)}\rightarrow v_i^{(\ell+1)}\).

 \subsection{Disaster-Conditioned Multi-Scale (DCMS) Kernel Prior}
\label{subsec:kernel}
Let \(p_i\in\R^2\) be the projected centroid of building \(i\), and \(d_{ij}=\|p_i-p_j\|_2\) the edge distance. We define a disaster-type embedding vector \(z_d\) (learned from a disaster-type token parsed from the xBD filename for event \(e\)). From \(z_d\), we predict the parameters of a multi-scale distance kernel:
\begin{align}
\pi(e) &= \mathrm{softmax}(W_\pi z_d + b_\pi),\\
\rho_m(e) &= \rho_{\min} + \mathrm{softplus}(w_m^\top z_d + b_m), \quad m=1,\dots,M,
\end{align}
where \(\pi(e)\) are nonnegative mixture weights that sum to one and determine the disaster-type-specific emphasis across short/medium/long ranges, and \(\rho_m(e)\) are disaster-type-specific length scales (shared across events of the same disaster type). Note that \(\pi(e)\) and \(\rho_m(e)\) are \emph{event-level} quantities in the sense that they are shared across all edges within event \(e\), while the resulting prior \(K_{ij}(e)\) remains \emph{edge-specific} through its dependence on \(d_{ij}\).

We use a simple exponential family basis:
\begin{equation}
\kappa_m(d_{ij}\mid e)=\exp\!\left(-\frac{d_{ij}}{\rho_m(e)+\varepsilon}\right),
\end{equation}
and define the DCMS spatial prior
\begin{equation}
K_{ij}(e)=\sum_{m=1}^{M} \pi_m(e)\,\kappa_m(d_{ij}\mid e),
\label{eq:kernel}
\end{equation}where $M=3$ captures \emph{short}, \emph{medium}, and \emph{long} range context with disaster-type-dependent emphasis.

\subsection{Kernel-Regularized Graph Attention}
\label{subsec:gat}
Given node features \(v_i^{(\ell)}\), fixed edge features \(u_{ij}\), and disaster-type embedding \(z_d\), we compute a compatibility score
\begin{equation}
\psi_{ij}^{(\ell)} = a_\phi\!\left([W_q v_i^{(\ell)}\,\|\,W_k v_j^{(\ell)}\,\|\,u_{ij}\,\|\,z_d]\right),
\end{equation}
where \(a_\phi(\cdot)\) is a Multi-Layer Perceptron (MLP). We then inject the kernel prior \eqref{eq:kernel} directly into the attention logit as a regularizer:
\begin{equation}
\ell_{ij}^{(\ell)}=\psi_{ij}^{(\ell)}+\lambda_K\log\big(K_{ij}(e)+\varepsilon\big).
\label{eq:attnlogit}
\end{equation}
Normalized attention is
\begin{equation}
\alpha_{ij}^{(\ell)}=\frac{\exp(\ell_{ij}^{(\ell)})}{\sum_{j'\in\mathcal{N}(i)} \exp(\ell_{ij'}^{(\ell)})}.
\end{equation}
This keeps attention data-adaptive while anchoring it to an interpretable, disaster-type-aware spatial prior. While the neighbourhood \(\mathcal{N}(i)\) is fixed by the kNN construction, the attention weights \(\alpha_{ij}^{(\ell)}\) are recomputed at each layer, yielding a dynamic, layer-dependent \emph{weighted} graph over the same edges. Equivalently, each layer induces weights on the fixed topology, i.e., \(\tilde G_e^{\ell}=(V_e,E_e,\{\alpha_{ij}^{(\ell)}\})\).

The graph update is standard message passing:
\begin{equation}
v_i^{(\ell+1)} = v_i^{(\ell)} + \sigma\!\left(\sum_{j\in\mathcal{N}(i)} \alpha_{ij}^{(\ell)}\,W_v v_j^{(\ell)}\right),
\end{equation}
followed by a classifier head producing class probabilities \(\hat y_i\in[0,1]^4\). Thus, during forward propagation the \emph{only} quantities updated by graph reasoning are the node representations \(\{v_i^{(\ell)}\}\) (and the induced attention weights \(\{\alpha_{ij}^{(\ell)}\}\)); the positions \(\{p_i\}\), geometric edge features \(\{u_{ij}\}\), and kNN topology \((V_e,E_e)\) remain fixed.

\subsection{Losses: Classification, Ordinal Consistency, and Residual De-correlation}
\label{subsec:loss}
We optimize a class-balanced classification loss with an ordinal auxiliary term and a residual Moran penalty:
\begin{equation}
\mathcal{L}=\mathcal{L}_{\text{CE}}+\lambda_{\text{emd}}\mathcal{L}_{\text{EMD}}+\lambda_{\text{res}}\mathcal{L}_{\text{res}}.
\end{equation}
For class imbalance, we use effective-number weights $w_c=\frac{1-\beta}{1-\beta^{n_c}}$ (normalized so $\sum_c w_c=4$) \cite{cui2019classbalanced}, giving
\begin{equation}
\mathcal{L}_{\text{CE}}=-\frac{1}{B}\sum_{i=1}^{B} w_{y_i}\log(\hat y_{i,y_i}+\varepsilon),
\end{equation}
where $\hat y_i\in[0,1]^4$ are predicted class probabilities \cite{lin2017focal}.

To encourage ordinally consistent errors over $\{0,1,2,3\}$, we use an EMD-style cumulative loss with $F_i^\star(t)=\sum_{c\le t}\mathbf{1}[y_i=c]$ and $\hat F_i(t)=\sum_{c\le t}\hat y_{ic}$:
\begin{equation}
\mathcal{L}_{\text{EMD}}=\frac{1}{B}\sum_{i=1}^{B}\sum_{t=0}^{3}\big(\hat F_i(t)-F_i^\star(t)\big)^2.
\end{equation}

Let severities be $s\in\{0,1,2,3\}$ and $\hat s_i=\sum_{c=0}^{3}c\,\hat y_{ic}$. For nodes from event $e$, define residuals $r_i=s_i-\hat s_i$ and centered residuals $\tilde r_i=r_i-\bar r$. With a fixed row-normalized spatial weight matrix $W^{(e)}$ on $(V_e,E_e)$, we compute a differentiable Moran surrogate
\begin{equation}
I_r^{(e)}=\frac{n_e}{\sum_{ij}W^{(e)}_{ij}}
\frac{\sum_{ij}W^{(e)}_{ij}\,\tilde r_i\tilde r_j}{\sum_i \tilde r_i^2+\varepsilon},
\label{eq:moran_resid}
\end{equation}
and penalize positive residual autocorrelation:
\begin{equation}
\mathcal{L}_{\text{res}}=\frac{1}{|\mathcal{E}_{\text{batch}}|}\sum_{e\in\mathcal{E}_{\text{batch}}}\max(0, I_r^{(e)}).
\end{equation}

\section{Experiments}
\label{sec:experiments}
\subsection{Implementation Details}
\label{subsec:impl}
This paper reports the measured values shown in the accompanying tables and figures. In our implementation, we use the xBD dataset (released for the xView2 challenge) with paired pre/post disaster RGB tiles (typically \(1024\times1024\), \(\sim 0.8\) m GSD), building polygons, and four building-level damage labels (no damage, minor, major, destroyed) \cite{gupta2019xbd,gupta2019creatingxbd,xview2dataset}. We crop a \ppc\ patch per polygon from the provided \texttt{xy} annotations (tight bounding box + small margin, then resize), compute centroids from the geographic annotations, project to local metric coordinates (UTM per event/AOI), and build a \(k\)-nearest-neighbour graph with \(k=16\) unless stated otherwise. The patch encoder is a ResNet-50 \cite{he2016resnet} with a bitemporal input stem, the kernel mixture uses \(M=3\) components, and optimization uses AdamW-style training \cite{kingma2015adam,loshchilov2019adamw}. We freeze the patch encoder and train only the graph and classifier heads in all graph variants and the patch-only baseline, to isolate context modelling. We report macro-F1 and per-class F1 for all experiments, and compute ordinal MAE on the severity scale \(\{0,1,2,3\}\). The loss weights \(\lambda_{\text{emd}}\) and \(\lambda_{\text{res}}\) were tuned to be 0.25 and 0.1 respectively in our experiments. For spatial behaviour, we report Moran's~I of \emph{residuals} (lower is better). For rollup-level reporting (e.g., \Cref{tab:dataset_spatial_rollup} and rollup diagnostics in LOEO), we group xBD events into \emph{disaster-type rollups} using the event token parsed from xBD filenames/metadata: \textit{Hurricanes} (Harvey, Florence, Michael, Matthew), \textit{Wildfires} (Santa Rosa, Carr, Woolsey, Pinery, Portugal), \textit{Floods} (Midwest U.S.\ floods; monsoon in Nepal/India/Bangladesh), \textit{Tornadoes} (Moore OK; Tuscaloosa AL; Joplin MO), \textit{Tsunamis} (Indonesia; Sunda Strait), \textit{Volcanoes} (Guatemala Fuego; Lower Puna), and \textit{Earthquakes} (Mexico City). This rollup mapping is used for analysis/aggregation, as well as for training and inference which also operates on per-event graphs with a disaster-type token as described in \Cref{subsec:kernel}.

\paragraph{Evaluation protocols.}
We use three protocols:
\begin{enumerate}
    \item \textbf{xView2 official holdout external reference.} We reproduce the official xView2 rows from the xView2 challenge website \cite{xview2challenge} and report our patch-only and graph variants based on their classification scores.

    \item \textbf{Zero-shot cross-event transfer (Leave-One-Event-Out; LOEO).} We hold out one event at a time and train on the remaining events. This is a harder stress test for cross-event generalization and spatial behaviour; it is \emph{not directly comparable} to the official holdout reference. In this protocol, the held-out event is evaluated with its associated \emph{disaster type} provided as metadata (consistent with the disaster-type-conditioned design in \Cref{subsec:kernel}). LOEO F1 scores are computed by pooling predictions over all held-out events and reporting global per-class F1 and macro-F1.

    \item \textbf{Zero-shot cross-dataset transfer (xBD$\rightarrow$Ida-BD).} We train on xBD and evaluate directly on Ida-BD \cite{lee2022idabd} with no Ida-BD labels used for training, providing only a disaster-type token at inference (Ida-BD is a hurricane event).
\end{enumerate}

\begin{figure*}[t]
    \centering
    \loadfig{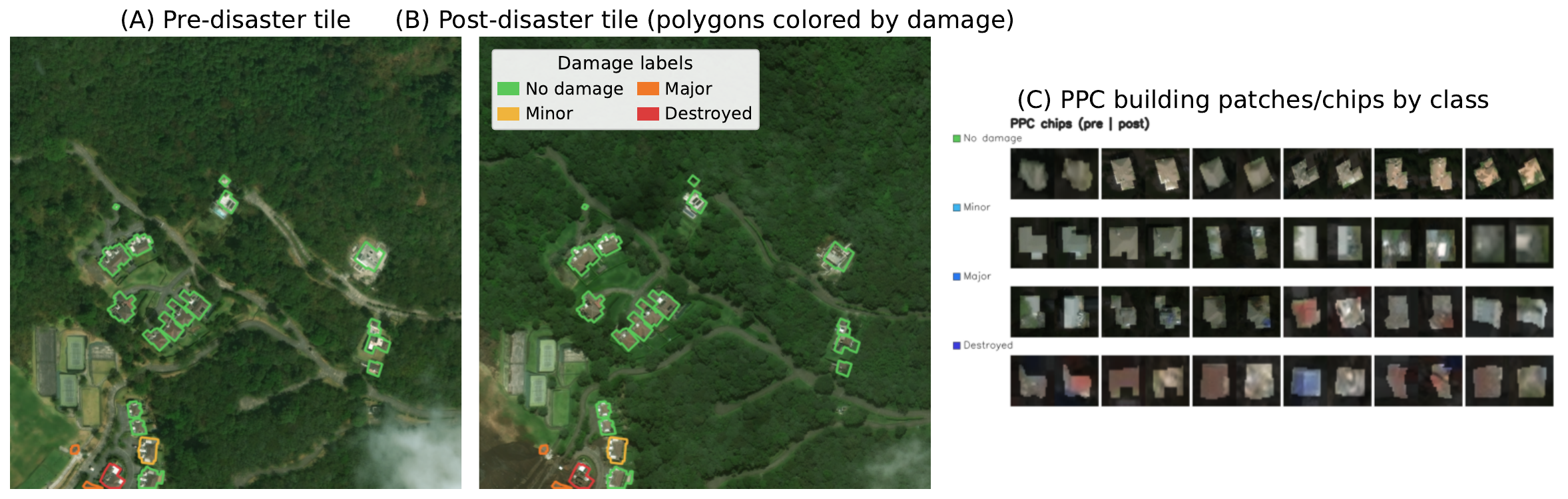}{width=0.99\textwidth}{2.3in}
    \caption{\ppc\ patch protocol: pre/post building-aligned crops are extracted from xBD polygons and resized for classification.}
    \label{fig:patchs_pca}
\end{figure*}

\subsection{Main Results: xView2 Holdout Comparison}
\label{subsec:holdout}
\Cref{tab:baseline_xview2_official_macro} provides a compact xView2 holdout reference. The official xView2 evaluation scores localization and classification jointly, whereas our protocol fixes building instances using provided polygons and evaluates the classification/context module only. Therefore, the official row should be read as a broad external reference point, not a direct head-to-head comparison. The meaningful comparison here is among the patch-only encoder, vanilla GAT, and our kernel-regularized graph model under the same fixed building instances and patch protocol. Under this controlled setting, Vanilla GAT improves Macro-F1 to \(0.84102\), and the full model further increases it to \(0.87288\), with gains concentrated in the harder \textit{Minor} and \textit{Major} classes. The full xView2 external-reference table is included in the supplementary material for completeness.

\begin{table}[t]
\centering
\small
\setlength{\tabcolsep}{4pt}
\caption{\textbf{xView2 holdout external reference.} Macro-F1 is the arithmetic mean of class F1 over U/Mi/Ma/D; official solution \#1 is shown only as a reference.}
\label{tab:baseline_xview2_official_macro}
\resizebox{\linewidth}{!}{%
\begin{tabular}{lcc}
\toprule
\textbf{Method} & \textbf{Macro-F1} & \textbf{Per-class F1 (U, Mi, Ma, D)} \\
\midrule
xView2 official solution \#1$^\dagger$ & 0.80446 & (0.92344, 0.64445, 0.78591, 0.86403) \\
Patch-only encoder & 0.82205 & (0.92880, 0.67320, 0.81520, 0.87100) \\
Vanilla GAT on patch encoder & 0.84102 & (0.93160, 0.70050, 0.84030, 0.89168) \\
\textbf{Ours (kernel + disaster + residual)} & \textbf{0.87288} & \textbf{(0.93400, 0.76200, 0.89200, 0.90352)} \\
\bottomrule
\end{tabular}%
}
\vspace{0.5mm}
{\footnotesize $^\dagger$ Official solution \#1 is shown only as an external reference and is not directly comparable to our fixed building instances and classification-only protocol.}
\end{table}

\subsection{Zero-shot Cross-Event Stress Test and Ablations (LOEO)}
\label{subsec:loeo}
The LOEO setting is a stricter test and better reflects event shift. \Cref{tab:ablation_loeo} shows three key patterns. First, the patch-only baseline is fragile under event shift (macro-F1 \(0.433\)). Second, vanilla GAT improves macro-F1 slightly (\(0.453\)) but leaves high residual Moran's~I (\(0.256\)) and can oversmooth mixed neighbourhoods. Third, the full model reaches the best LOEO macro-F1 (\(0.503\)) while dramatically lowering residual Moran's~I to \(0.079\), which is the main empirical signature of the proposed objective.

Among single components, the kernel prior contributes the strongest standalone gain in both macro-F1 and spatial consistency. Disaster conditioning alone yields a smaller but consistent improvement, and the residual term alone mostly attacks spatially structured error (large Moran drop, smaller F1 gain). The pairwise ablations support the same narrative: \textit{kernel + residual} is stronger than either component alone, and adding disaster conditioning produces the best final result. \Cref{tab:compact_sensitivity} adds two sensitivity checks: bitemporal \ppc\ input improves the difficult Minor/Major distinction over post-only imagery, and the full model is stable around \(k=16\), which slightly minimizes residual Moran's~I among the tested graph connectivities.

\begin{table*}[t]
\centering
\small
\setlength{\tabcolsep}{4pt}
\caption{\textbf{LOEO ablation study.} Residual Moran's~I is computed on severity residuals (lower is better).}
\label{tab:ablation_loeo}
\resizebox{\textwidth}{!}{%
\begin{tabular}{l c c c c c c c}
\toprule
\textbf{Variant} & \textbf{Kernel} & \textbf{Disaster} & \textbf{Residual} & \textbf{LOEO} & \textbf{Per-class F1} & \textbf{Residual} & \textbf{Severity} \\
& \textbf{prior} & \textbf{cond.} & \textbf{loss} & \textbf{Macro-F1} & \textbf{(No, Mi, Ma, D)} & \textbf{Moran's I $\downarrow$} & \textbf{MAE $\downarrow$} \\
\midrule
Patch-only (xView2 1st-place-style Res50 backend) & $\times$ & $\times$ & $\times$ & 0.433 & (0.61, 0.37, 0.41, 0.34) & 0.248 & 0.562 \\
Vanilla GAT (kNN graph baseline) & $\times$ & $\times$ & $\times$ & 0.453 & (0.57, 0.40, 0.44, 0.40) & 0.256 & 0.541 \\
+ Kernel prior only & $\checkmark$ & $\times$ & $\times$ & 0.473 & (0.60, 0.42, 0.46, 0.41) & 0.182 & 0.507 \\
+ Disaster conditioning only & $\times$ & $\checkmark$ & $\times$ & 0.460 & (0.59, 0.41, 0.45, 0.39) & 0.205 & 0.521 \\
+ Residual de-correlation only & $\times$ & $\times$ & $\checkmark$ & 0.450 & (0.58, 0.39, 0.44, 0.39) & 0.139 & 0.529 \\
+ Kernel prior + Disaster conditioning & $\checkmark$ & $\checkmark$ & $\times$ & 0.488 & (0.61, 0.44, 0.48, 0.42) & 0.116 & 0.484 \\
+ Kernel prior + Residual loss & $\checkmark$ & $\times$ & $\checkmark$ & 0.490 & (0.60, 0.44, 0.49, 0.43) & 0.096 & 0.474 \\
\textbf{Full (Kernel + Disaster + Residual)} & $\checkmark$ & $\checkmark$ & $\checkmark$ & \textbf{0.503} & \textbf{(0.62, 0.45, 0.50, 0.44)} & \textbf{0.079} & \textbf{0.458} \\
\bottomrule
\end{tabular}%
}
\end{table*}

\begin{table}[t]
\centering
\scriptsize
\setlength{\tabcolsep}{4pt}
\renewcommand{\arraystretch}{0.95}
\caption{\textbf{Sensitivity checks.} Input sensitivity uses the fixed-instance holdout protocol; $k$-sensitivity uses LOEO. Mi/Ma averages Minor and Major F1.}
\label{tab:compact_sensitivity}
\begin{tabular*}{\linewidth}{@{\extracolsep{\fill}}ccc lcc@{}}
\toprule
\multicolumn{3}{c}{\textbf{Graph sensitivity}} &
\multicolumn{3}{c}{\textbf{Input sensitivity}} \\
\midrule
\textbf{$k$} & \textbf{Macro-F1} & \textbf{Moran's I} &
\textbf{Setting} & \textbf{Macro-F1} & \textbf{Mi/Ma F1} \\
\midrule
8  & 0.498 & 0.081 & Post-only & 0.828 & 0.762 \\
16 & \textbf{0.503} & \textbf{0.079} & Pre+Post (\ppc) & \textbf{0.873} & \textbf{0.827} \\
32 & 0.500 & 0.084 & & & \\
\bottomrule
\end{tabular*}
\end{table}

\begin{figure*}[t]
    \centering
    \loadfig{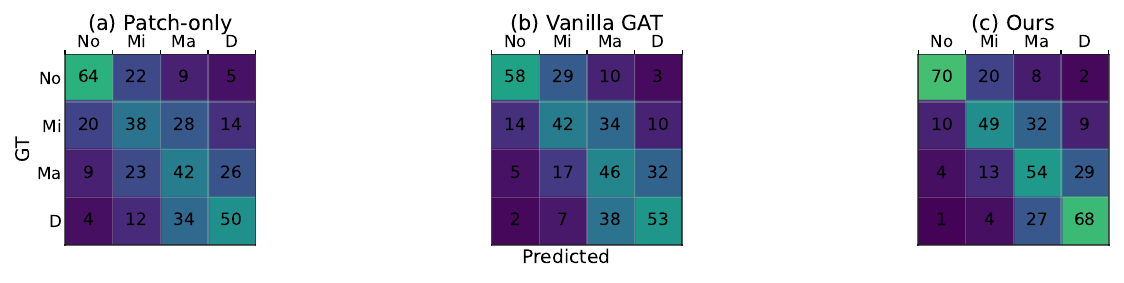}{width=0.99\textwidth}{3.0in}
     \caption{\textbf{LOEO confusion matrices.} Row-normalized percentages under the LOEO protocol.}
     
    \label{fig:conf_mat}
\end{figure*}

\Cref{fig:conf_mat} makes the LOEO error modes more explicit. Patch-only produces broader off-diagonal mass under event shift, especially across adjacent severity levels and occasional severe confusions. Vanilla GAT improves some local consistency, but its confusion mass shifts toward adjacent-class transitions consistent with oversmoothing-like behaviour (e.g., stronger No$\rightarrow$Minor and Destroyed$\rightarrow$Major leakage in mixed neighbourhoods). The proposed model increases diagonal concentration across all classes while keeping the realistic hard confusions concentrated in \textit{Minor}$\leftrightarrow$\textit{Major} and \textit{Major}$\leftrightarrow$\textit{Destroyed}, which matches the qualitative boundary behaviour outlined later in \Cref{fig:qual_maps}.

\vspace{1mm}

\subsection{Zero-shot Cross-Dataset Transfer: xBD $\rightarrow$ Ida-BD}
\label{subsec:idabd}
To probe generalization beyond xBD, we evaluate zero-shot transfer from xBD to Ida-BD, a high-resolution satellite dataset constructed for Hurricane Ida with paired pre/post tiles and polygon-level building annotations \cite{lee2022idabd}. \Cref{tab:zeroshot_idabd} reports published xBD$\rightarrow$Ida-BD transfer baselines and our patch-based results when training only on xBD and evaluating directly on Ida-BD while providing a disaster-type token at inference. Prior work reports transfer performance in end-to-end localization+classification or segmentation-style pipelines (and may merge classes such as Destroyed into Major), which makes their reported scores not directly comparable to our polygon-patch classification protocol. We therefore include these reported baselines primarily to contextualize the difficulty of the dataset shift, and emphasize the relative gains among the patch-based baselines and our model under a consistent setup.

\begin{table*}[t]
\centering
\small
\setlength{\tabcolsep}{4pt}
\caption{\textbf{Zero-shot transfer from xBD to Ida-BD.} Models train on xBD and evaluate directly on Ida-BD.}
\label{tab:zeroshot_idabd}
\resizebox{\textwidth}{!}{%
\begin{tabular}{l c c c c}
\toprule
\textbf{Method} & \textbf{Pipeline / notes} & \textbf{Token} & \textbf{Macro-F1} & \textbf{Per-class F1 (No, Mi, Ma, D)} \\
\midrule
Siam-UNet  \cite{kaur2023dahitra} & D merged into Ma$^\dagger$ & -- & 0.458 & (0.916, 0.208, 0.251, --) \\
xView2 winner  \cite{parupati2025robust} & per-class weighted F1 & -- & 0.268 & (0.667, 0.211, 0.154, 0.041) \\
\midrule
Patch-only (ours) & patch cls (polygons) & \checkmark & 0.275 & (0.670, 0.190, 0.160, 0.080) \\
Vanilla GAT (ours) & patch cls + graph & \checkmark & 0.295 & (0.690, 0.220, 0.180, 0.090) \\
\textbf{Ours (Kernel + Disaster + Residual)} & patch cls + graph & \checkmark & \textbf{0.335} & \textbf{(0.710, 0.270, 0.230, 0.130)} \\
\bottomrule
\end{tabular}%
}
\vspace{1mm}

{\footnotesize $^\dagger$ DAHiTrA reports three damage classes on Ida-BD (Destroyed merged into Major); Macro-F1 is averaged over the reported damage classes.}
\end{table*}

\subsection{Qualitative and Spatial Diagnostics}
\label{subsec:qual}
\begin{figure*}[t]
\centering
\begin{subfigure}[t]{0.34\linewidth}
\centering
\loadfig{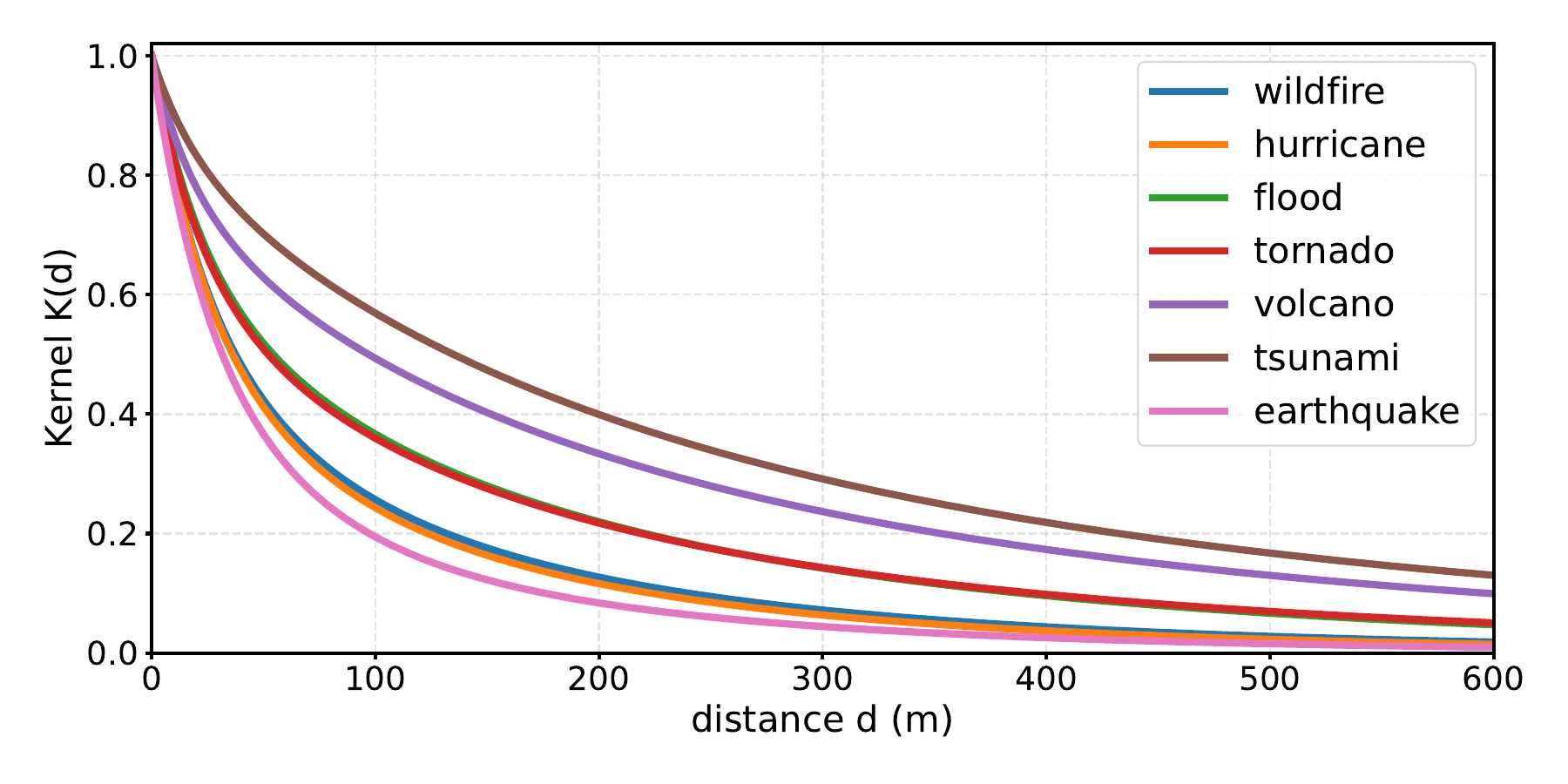}{width=\linewidth}{2.0in}
\caption{Learned disaster-type-conditioned kernel priors by rollup.}
\label{fig:kernel_overlay}
\end{subfigure}%
\hfill
\begin{subfigure}[t]{0.64\linewidth}
\centering
\loadfig{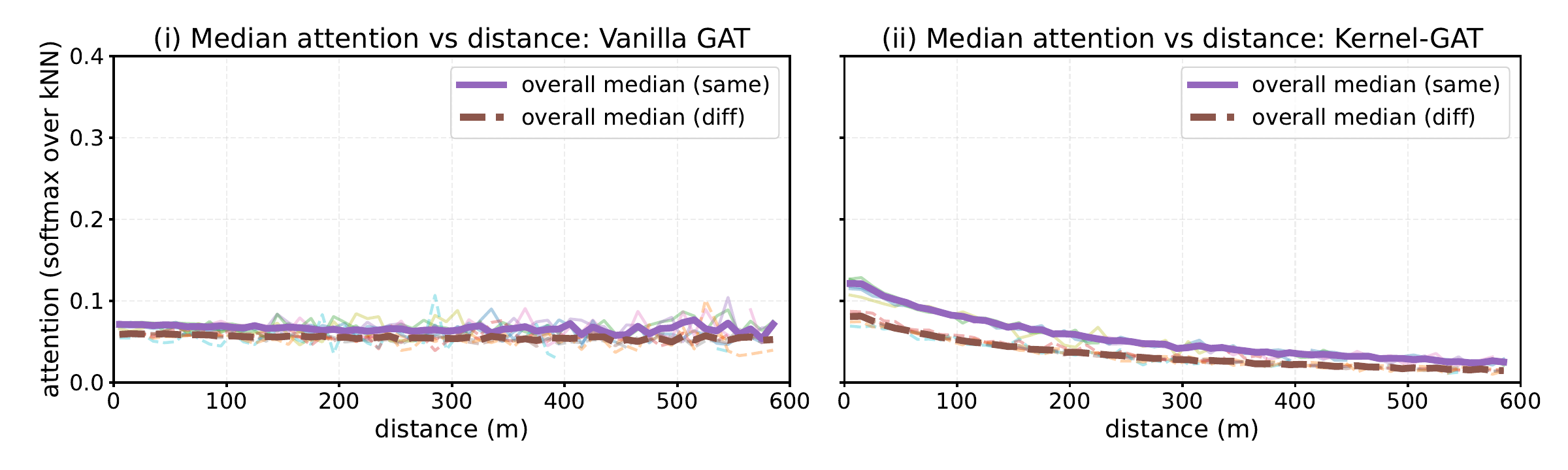}{width=\linewidth}{2.0in}
\caption{Attention--distance diagnostics; solid/dashed curves denote same/different-label edges.}
\label{fig:attn_overlay}
\end{subfigure}
\caption{\textbf{Kernel and attention-distance diagnostics.}
Full rollup-specific grids and kernel parameter tables are provided in the supplementary material.}
\label{fig:kernel_attention_diagnostics}
\end{figure*}

\Cref{fig:kernel_overlay} visualizes the learned mean disaster-type-conditioned kernel priors aggregated by disaster rollup. Beyond decaying with distance, the key result is that the \emph{mixture shape} changes: some rollups emphasize shorter ranges while others assign more weight to medium/long scales. This pattern shows qualitative consistency with the spatial statistics in \Cref{tab:dataset_spatial_rollup}: for rollups with slower spatial decay, more distant buildings can remain informative because damage labels are more spatially correlated. This supports our core motivation that ``useful neighbourhood size'' is disaster-type dependent.

\Cref{fig:attn_overlay} compares how attention weights vary with inter-building distance for vanilla GAT and the proposed kernel-regularized GAT. The plotted medians are computed separately for edges linking buildings with the same damage label and edges linking buildings with different damage labels. In vanilla GAT, attention is less consistently structured with distance and shows weaker separation between same-label and different-label edges, reflecting noisier neighbourhood aggregation. In contrast, our kernel-regularized attention follows a more coherent distance-decay trend while preserving feature-driven variation, which reduces pathological long-range links and stabilizes local smoothing. Additional rollup-level plots and per-rollup kernel tables are provided in the supplementary material. 

\Cref{fig:qual_maps} shows a representative LOEO evaluation tile (Hurricane-Harvey), showcased due to the tile's rich spatial autocorrelation (high Moran's I). The patch-only model, which was not trained on Hurricane Harvey, becomes sensitive to pre/post appearance shifts and produces mild spatially autocorrelated residuals because its errors are presumably driven by latent scene factors (hazard intensity, neighbourhood morphology, and imaging conditions) that are themselves spatially clustered, even without any explicit spatial message passing. Vanilla GAT improves local coherence but exhibits oversmoothing-like errors in mixed neighbourhoods. Our model is more stable along boundaries and maintains a cleaner damage pattern without collapsing mixed regions, even though some oversmoothing errors seem to be present.

\Cref{tab:residual_moran} summarizes rollup-level LOEO Macro-F1 and residual Moran's~I in a compact F1/Moran format. The proposed model generally improves rollup-level accuracy while consistently lowering residual autocorrelation relative to both baselines, matching the intended effect of the residual de-correlation objective. 

\begin{figure*}[t]
    \centering
    \loadfig{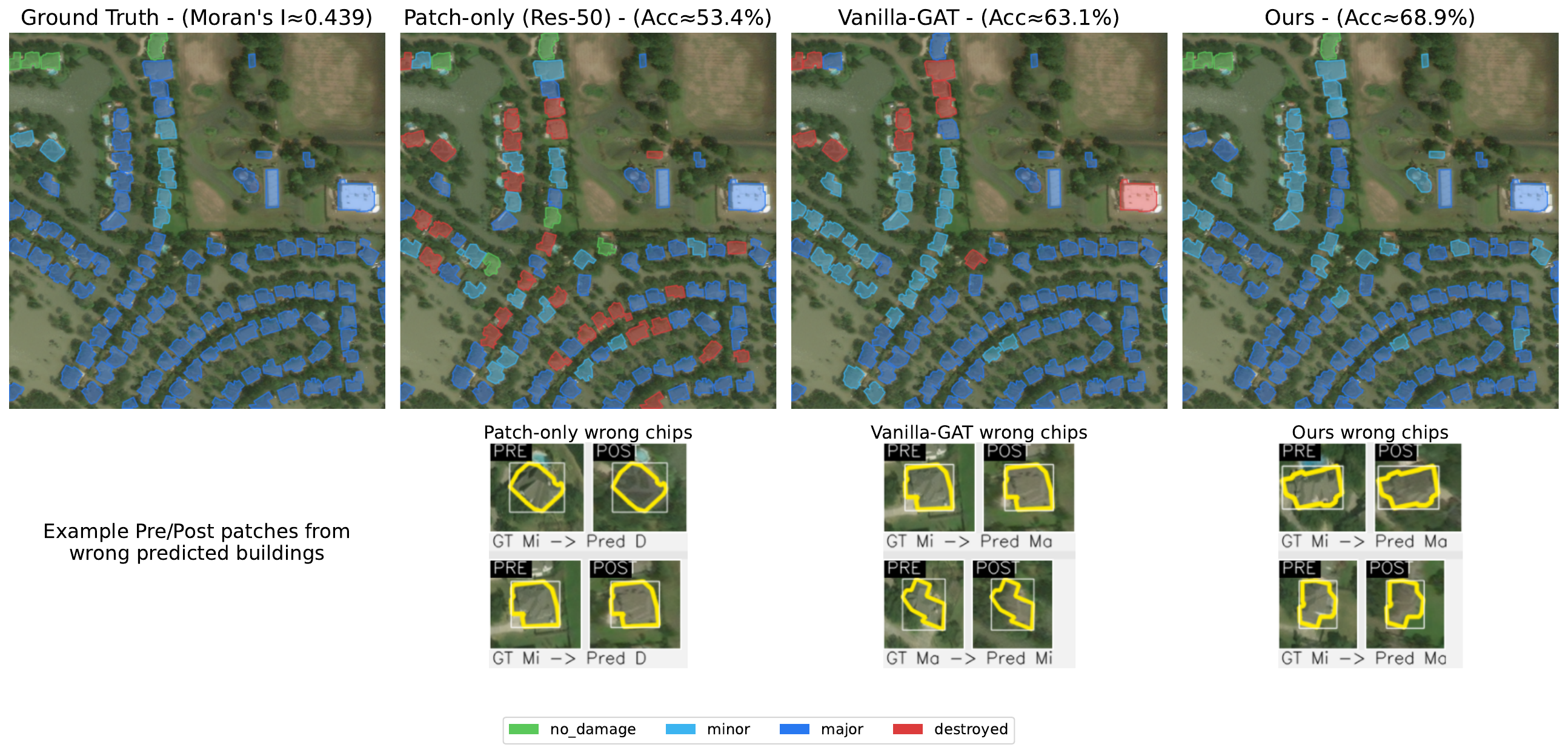}{width=0.99\textwidth}{3.0in}
 \caption{LOEO qualitative map on a high-Moran Hurricane-Harvey tile. Left-to-right: ground truth, patch-only, vanilla GAT, and ours.}
    \label{fig:qual_maps}
\end{figure*}

\begin{table}[t]
\centering
\small
\setlength{\tabcolsep}{4pt}
\renewcommand{\arraystretch}{1.05}
\caption{\textbf{Rollup LOEO Macro-F1 / residual Moran's~I.} Higher F1 and lower Moran's~I are better.}
\label{tab:residual_moran}
\resizebox{\linewidth}{!}{%
\begin{tabular}{lccccccc}
\toprule
\textbf{Method} & \textbf{earthquake} & \textbf{flood} & \textbf{hurricane} & \textbf{tornado} & \textbf{tsunami} & \textbf{volcano} & \textbf{wildfire} \\
\midrule
Patch-only & .470/.160 & .397/.313 & .438/.263 & .424/.368 & .415/.427 & .382/.359 & .505/.260 \\
Vanilla GAT & .482/.174 & .416/.335 & .459/.283 & .442/.384 & .429/.437 & .401/.364 & .542/.283 \\
\textbf{Ours} & \textbf{.548/.134} & \textbf{.474/.212} & \textbf{.517/.198} & \textbf{.502/.294} & \textbf{.488/.301} & \textbf{.445/.320} & \textbf{.547/.204} \\
\bottomrule
\end{tabular}%
}
\end{table}
\section{Discussion}
\label{sec:discussion}

The holdout and LOEO results serve different purposes. The xView2 holdout numbers provide an external reference scale for damage-classification performance, but the controlled comparisons are the patch-only, vanilla GAT, and proposed graph variants under fixed building instances. The post-only check confirms that bitemporal \ppc\ input remains useful even in this building-level setting, especially for the Minor/Major distinction where pre-disaster appearance reduces ambiguity. LOEO is the more diagnostic setting for our claim: when the held-out event changes roof materials, imaging conditions, or disaster morphology, patch-only predictions leave spatially clustered residuals, indicating that appearance errors are not independent across space. This is also why we view the fixed-instance protocol as a feature rather than a shortcut: by holding localization constant, the experiments isolate whether spatial context itself improves damage classification under event shift.

Vanilla GAT confirms that graph context is useful but not sufficient. It improves over the patch-only encoder, yet unconstrained attention can still propagate mistakes across mixed neighbourhoods, especially near damage boundaries. The issue is not graph reasoning itself; it is the lack of a principled, disaster-type-specific spatial scale. The learned DCMS kernels address this by making the effective context radius conditional on disaster type, while the residual Moran penalty discourages coherent-but-wrong neighbourhoods; the compact (k)-sensitivity check shows this behaviour is stable around the chosen (k=16). More broadly, the results suggest that spatial context should be treated as a conditional cue rather than a universal prior: when labels are strongly clustered, neighbouring buildings can provide useful evidence, but when spatial clustering is weak, forcing neighbourhood agreement can become a source of bias.

The spatial diagnostics support this interpretation: the learned kernels do not collapse to a single default curve, and \Cref{tab:residual_moran} shows the F1/Moran trade-off across disaster rollups rather than only in aggregate. The cross-dataset transfer results in \Cref{tab:zeroshot_idabd} provide an additional stress test under xBD$\rightarrow$Ida-BD dataset shift. Although published Ida-BD baselines use different pipelines and label handling \cite{kaur2023dahitra,parupati2025robust}, the fixed building instances and patch protocol show the same trend: the proposed model improves over patch-only and vanilla GAT, with gains concentrated in the harder \textit{Minor} and \textit{Major} classes. This does not imply that instance-level graph reasoning replaces full-image models; rather, it addresses a complementary question. Full-image systems jointly solve localization, scene understanding, and classification, whereas our controlled setting asks whether, once candidate buildings are available, relational spatial structure can improve the robustness of building-level damage labels.

Some notable limitations remain. First, the method is not an end-to-end xView2 system: it assumes candidate building instances and centroids are available, and our experiments use the provided xBD polygons to avoid confounding localization quality with contextual reasoning. In deployment, these instances could come from a detector, segmenter, cadastral layer, or OSM-style footprint source, but errors in those inputs would affect crop quality, node availability, and graph geometry. Evaluating the method with detector-generated footprints and explicitly modelling localization uncertainty are therefore important next steps. Second, our GPS-only \(k\)-NN graphs ignore scene semantics (e.g., roads, parcels, barriers) and can connect buildings that are near in Euclidean space but weakly coupled in damage. Third, we assume the disaster-type token is known at inference; robustness to missing, noisy, or ambiguous event metadata is untested. Finally, we use optical \ppc\ patches only and do not leverage modalities such as SAR that matter under cloud/smoke. Future work can address these by using topology-/uncertainty-aware edges, inferring conditioning when metadata is unavailable, integrating detector-generated instances, and extending to multimodal inputs while retaining residual spatial diagnostics.

\section{Conclusion}
\label{sec:conclusion}
We presented a disaster-type-conditioned, kernel-regularized graph attention model for xBD building damage classification in the controlled post-localization, classification-only xBD setting using a clean \ppc\ building-patch protocol. The method keeps local evidence ``close'' by retaining strong per-building patch representations, but improves reliability by explicitly modelling spatial behaviour instead of relying on naive image-only recognition under shift. It then brings the \emph{right} neighbours closer by injecting a disaster-type-adaptive, multi-scale geostatistical prior into attention, so neighbourhood influence is distance- and type-aware rather than an unconstrained tendency to smooth. A residual Moran objective further discourages coherent-but-wrong neighbourhood errors by penalizing spatial autocorrelation in severity residuals. Across an external-reference comparison and zero-shot transfer evaluations under event shift (xBD LOEO) and dataset shift (xBD$\rightarrow$Ida-BD), the model improves macro-F1 and reduces residual Moran's~I, supporting the central claim: better event generalization starts with keeping your ``friends'' close (i.e., learning to utilize the rich spatial relationships in disaster damage patterns to make more informed predictions), and the \emph{right} neighbours closer (i.e., also learning \emph{which} spatial context to trust and at what scale, not by smoothing more aggressively).

\section*{Acknowledgements}

This work was supported by the Natural Sciences and Engineering Research Council of Canada (NSERC) Alliance Program (ALLRP 570826-21), in partnership with Statistics Canada, Natural Resources Canada, Credit Valley
Conservation, Toronto and Region Conservation Authority, and the City of Kitchener.

The authors acknowledge the use of OpenAI's ChatGPT for manuscript preparation, including drafting, grammar and structure refinement, and assistance in identifying potentially relevant references. All scientific content, results, interpretations, and cited references were manually reviewed and verified by the authors.

\bibliographystyle{splncs04}
\bibliography{refs}

@misc{gupta2019creatingxbd,
author       = {Gupta, Ritwik and Goodman, Bryce and Patel, Nirav and Hosfelt, Richard and Sajeev, Sandra and Heim, Eric and Doshi, Jigar and Lucas, Keane and Choset, Howard and Gaston, Matthew},
title        = {Creating {xBD}: A Dataset for Assessing Building Damage from Satellite Imagery},
year         = {2019},
month        = may,
publisher    = {Carnegie Mellon University},
howpublished = {KiltHub preprint},
doi          = {10.1184/R1/8135576.v1},
note         = {Accessed: 2026-06-26}
}

@article{gupta2019xbd,
  author  = {Ritwik Gupta and Richard Hosfelt and Sandra Sajeev and Nirav Patel and Bryce Goodman and Jigar Doshi and Eric Heim and Howie Choset and Matthew Gaston},
  title   = {x{BD}: A Dataset for Assessing Building Damage from Satellite Imagery},
  journal = {arXiv preprint arXiv:1911.09296},
  year    = {2019},
  doi     = {10.48550/arXiv.1911.09296}
}

@misc{xview2dataset,
  author       = {{xView2 Challenge Organizers}},
  title        = {x{View2} Dataset},
  howpublished = {\url{https://xview2.org/dataset}},
  note         = {Accessed: 2026-06-26}
}

@article{weber2020building,
  author  = {Ethan Weber and Hassan Kan{\'e}},
  title   = {Building Disaster Damage Assessment in Satellite Imagery with Multi-Temporal Fusion},
  journal = {arXiv preprint arXiv:2004.05525},
  year    = {2020},
  doi     = {10.48550/arXiv.2004.05525}
  
}

@article{shen2022bdanet,
  author  = {Yu Shen and Sijie Zhu and Taojiannan Yang and Chen Chen and Delu Pan and Jianyu Chen and Liang Xiao and Qian Du},
  title   = {{BDANet}: Multiscale Convolutional Neural Network with Cross-Directional Attention for Building Damage Assessment from Satellite Images},
  journal = {IEEE Transactions on Geoscience and Remote Sensing},
  year    = {2022},
  volume  = {60},
  pages   = {1--14},
  doi     = {10.1109/TGRS.2021.3080580}
}

@article{deng2022improvedunet,
  author  = {Liwei Deng and Yue Wang},
  title   = {Post-Disaster Building Damage Assessment Based on Improved {U-Net}},
  journal = {Scientific Reports},
  year    = {2022},
  volume  = {12},
  pages   = {15862},
  doi     = {10.1038/s41598-022-20114-w}
}

@inproceedings{velivckovic2018gat,
  author    = {Petar Veli{\v{c}}kovi{\'c} and Guillem Cucurull and Arantxa Casanova and Adriana Romero and Pietro Li{\`o} and Yoshua Bengio},
  title     = {Graph Attention Networks},
  booktitle = {International Conference on Learning Representations (ICLR)},
  year      = {2018},
  url       = {https://openreview.net/forum?id=rJXMpikCZ},
    note         = {Accessed: 2026-06-26}
}

@inproceedings{brody2022gatv2,
  author    = {Shaked Brody and Uri Alon and Eran Yahav},
  title     = {How Attentive Are Graph Attention Networks?},
  booktitle = {International Conference on Learning Representations (ICLR)},
  year      = {2022},
  url       = {https://openreview.net/forum?id=F72ximsx7C1},
    note         = {Accessed: 2026-06-26}
}

@article{battaglia2018relational,
  author  = {Peter W. Battaglia and Jessica B. Hamrick and Victor Bapst and Alvaro Sanchez-Gonzalez and Vin{\'\i}cius Zambaldi and Mateusz Malinowski and Andrea Tacchetti and David Raposo and Adam Santoro and Ryan Faulkner and {\c{C}}aglar G{\"u}l{\c{c}}ehre and Francis Song and Andrew Ballard and Justin Gilmer and George E. Dahl and Ashish Vaswani and Kelsey Allen and Charles Nash and Victoria Langston and Chris Dyer and Nicolas Heess and Daan Wierstra and Pushmeet Kohli and Matthew Botvinick and Oriol Vinyals and Yujia Li and Razvan Pascanu},
  title   = {Relational Inductive Biases, Deep Learning, and Graph Networks},
  journal = {arXiv preprint arXiv:1806.01261},
  year    = {2018},
  doi     = {10.48550/arXiv.1806.01261},
    note         = {Accessed: 2026-06-26}
}

@article{tobler1970computer,
  author  = {Waldo R. Tobler},
  title   = {A Computer Movie Simulating Urban Growth in the Detroit Region},
  journal = {Economic Geography},
  year    = {1970},
  volume  = {46},
  number  = {sup1},
  pages   = {234--240},
  doi     = {10.2307/143141}
}

@book{cressie1993statistics,
  author    = {Noel A. C. Cressie},
  title     = {Statistics for Spatial Data},
  publisher = {Wiley},
  year      = {1993},
  edition   = {Revised},
  doi       = {10.1002/9781119115151}
}

@article{moran1950notes,
  author  = {P. A. P. Moran},
  title   = {Notes on Continuous Stochastic Phenomena},
  journal = {Biometrika},
  year    = {1950},
  volume  = {37},
  number  = {1/2},
  pages   = {17--23},
  doi     = {10.2307/2332142}

}

@misc{xview2challenge,
  author       = {{xView2 Challenge Organizers}},
  title        = {x{View2} Challenge Leaderboard},
  howpublished = {\url{https://xview2.org/challenge}},
  note         = {Accessed: 2026-06-26}
}

@inproceedings{hamilton2017inductive,
  author    = {William L. Hamilton and Rex Ying and Jure Leskovec},
  title     = {Inductive Representation Learning on Large Graphs},
  booktitle = {Advances in Neural Information Processing Systems (NeurIPS)},
  year      = {2017},
  volume    = {30},
  doi = {https://doi.org/10.48550/arXiv.1706.02216}
}

@inproceedings{kipf2017semi,
  author    = {Thomas N. Kipf and Max Welling},
  title     = {Semi-Supervised Classification with Graph Convolutional Networks},
  booktitle = {International Conference on Learning Representations (ICLR)},
  year      = {2017},
  url       = {https://openreview.net/forum?id=SJU4ayYgl},
    note         = {Accessed: 2026-06-26}
}

@misc{hasan2025uavgat,
  author       = {Fuad Hasan and Ali Lesani and Chul Min Yeum and Rodrigo Costa},
  title        = {Graph-Attention Network for Spatially-Aware Post-Hurricane Building Damage Assessment from {UAV} Imagery},
  howpublished = {ISPRS Congress},
  year         = {2026},
  note         = {Accepted for presentation},
    note         = {Accessed: 2026-06-26}
}

@article{geary1954contiguity,
  author  = {R. C. Geary},
  title   = {The Contiguity Ratio and Statistical Mapping},
  journal = {The Incorporated Statistician},
  year    = {1954},
  volume  = {5},
  number  = {3},
  pages   = {115--146},
  doi     = {10.2307/2986645}

}

@inproceedings{he2016resnet,
  author    = {Kaiming He and Xiangyu Zhang and Shaoqing Ren and Jian Sun},
  title     = {Deep Residual Learning for Image Recognition},
  booktitle = {Proceedings of the IEEE Conference on Computer Vision and Pattern Recognition (CVPR)},
  year      = {2016},
  pages     = {770--778},
  doi       = {10.1109/CVPR.2016.90}
}

@inproceedings{cui2019classbalanced,
  author    = {Yin Cui and Menglin Jia and Tsung-Yi Lin and Yang Song and Serge Belongie},
  title     = {Class-Balanced Loss Based on Effective Number of Samples},
  booktitle = {Proceedings of the IEEE/CVF Conference on Computer Vision and Pattern Recognition (CVPR)},
  year      = {2019},
  pages     = {9268--9277},
  doi       = {10.1109/CVPR.2019.00949}
}

@inproceedings{lin2017focal,
  author    = {Tsung-Yi Lin and Priya Goyal and Ross Girshick and Kaiming He and Piotr Doll{\'a}r},
  title     = {Focal Loss for Dense Object Detection},
  booktitle = {Proceedings of the IEEE International Conference on Computer Vision (ICCV)},
  year      = {2017},
  pages     = {2980--2988},
  doi       = {10.1109/ICCV.2017.324}
}

@inproceedings{kingma2015adam,
  author    = {Diederik P. Kingma and Jimmy Ba},
  title     = {Adam: A Method for Stochastic Optimization},
  booktitle = {International Conference on Learning Representations (ICLR)},
  year      = {2015},
  url       = {https://arxiv.org/abs/1412.6980},
    note         = {Accessed: 2026-06-26}
}

@inproceedings{loshchilov2019adamw,
  author    = {Ilya Loshchilov and Frank Hutter},
  title     = {Decoupled Weight Decay Regularization},
  booktitle = {International Conference on Learning Representations (ICLR)},
  year      = {2019},
  url       = {https://openreview.net/forum?id=Bkg6RiCqY7},
      note         = {Accessed: 2026-06-26}
}

@misc{lee2022idabd,
  author       = {Cheng{-}Chun Lee and Navjot Kaur and Ali Mahdavi{-}Amiri and Ali Mostafavi},
  title        = {{Ida-BD}: Pre- and Post-Disaster High-Resolution Satellite Imagery for Building Damage Assessment from Hurricane Ida},
  howpublished = {DesignSafe-CI dataset release},
  year         = {2022},
  url          = {https://www.designsafe-ci.org/data/browser/public/designsafe.storage.published/PRJ-3563},
    note         = {Accessed: 2026-06-26}
}

@article{kaur2023dahitra,
  author  = {Navjot Kaur and Cheng{-}Chun Lee and Ali Mostafavi and Ali Mahdavi{-}Amiri},
  title   = {Large-Scale Building Damage Assessment Using a Novel Hierarchical Transformer Architecture on Satellite Images},
  journal = {Computer-Aided Civil and Infrastructure Engineering},
  volume  = {38},
  number  = {15},
  pages   = {2072--2091},
  year    = {2023},
  doi     = {10.1111/mice.12981}
}

@inproceedings{parupati2025robust,
  author    = {Bharath Chandra Reddy Parupati and Shruti Kshirsagar and Rajiv Bagai and Atri Dutta},
  title     = {Towards Robust Building Damage Detection: Leveraging Augmentation and Domain Adaptation},
  booktitle = {2025 IEEE Green Technologies Conference (GreenTech)},
  year      = {2025},
  pages     = {163--167},
  doi       = {10.1109/GreenTech62170.2025.10977720}
}

@article{bright2025dataset,
  author  = {Hongruixuan Chen and Jian Song and Olivier Dietrich and Clifford Broni{-}Bediako and Weihao Xuan and Junjue Wang and Xinlei Shao and Yimin Wei and Junshi Xia and Cuiling Lan and Konrad Schindler and Naoto Yokoya},
  title   = {Bright: A Globally Distributed Multimodal Building Damage Assessment Dataset with Very-High-Resolution for All-Weather Disaster Response},
  journal = {Earth System Science Data},
  year    = {2025},
  volume  = {17},
  pages   = {6217--6253},
  doi     = {10.5194/essd-17-6217-2025}
}

@article{melamed2023xfbd,
  author  = {Dennis Melamed and Cameron Johnson and Chen Zhao and Russell Blue and Philip Morrone and Anthony Hoogs and Brian Clipp},
  title   = {{xFBD}: Focused Building Damage Dataset and Analysis},
  journal = {arXiv preprint arXiv:2212.13876},
  year    = {2023},
  doi     = {10.48550/arXiv.2212.13876}
}

\end{document}